\documentclass[sigconf]{acmart}

\usepackage{multirow}
\usepackage{pifont}
\usepackage{enumitem}
\usepackage{wrapfig}
\usepackage{subfig}
\usepackage{graphicx}
\usepackage{natbib}
\AtBeginDocument{%
  }

\setcopyright{acmlicensed}
\copyrightyear{2018}
\acmYear{2018}
\acmDOI{XXXXXXX.XXXXXXX}
\acmConference[Conference acronym 'XX]{Make sure to enter the correct
  conference title from your rights confirmation email}{June 03--05,
  2018}{Woodstock, NY}
\acmISBN{978-1-4503-XXXX-X/2018/06}

\begin{document}

\title{One Model for One Graph: A New Perspective for Pretraining with Cross-dataset Graphs}

\author{Jingzhe Liu}
\affiliation{%
  \institution{Michigan State University}
  \country{USA}
}

\author{Haitao Mao}
\affiliation{%
  \institution{Michigan State University}
  \country{USA}}

\author{Zhikai Chen}
\affiliation{%
  \institution{Michigan State University}
  \country{USA}}

\author{Bingheng Li}
\affiliation{%
  \institution{Michigan State University}
  \country{USA}}

\author{Wenqi Fan}
\affiliation{%
  \institution{Hong Kong Polytechnic University}
  \city{Hong Kong}
  \country{China}}

\author{Mingxuan Ju}
\affiliation{%
  \institution{Snap Inc.}
  \country{USA}}

\author{Mingxuan Ju}
\affiliation{%
  \institution{Snap Inc.}
  \country{USA}}

\author{Tong Zhao}
\affiliation{%
  \institution{Snap Inc.}
  \country{USA}}  

\author{Neil Shah}
\affiliation{%
  \institution{Snap Inc.}
  \country{USA}}

\author{Jiliang Tang}
\affiliation{%
  \institution{Michigan State University}
  \country{USA}}

\renewcommand{\shortauthors}{Trovato et al.}
\newcommand\liu[1]{\textcolor{blue}{liu: #1}}
\begin{abstract}
Graph Neural Networks (GNNs) have emerged as a powerful tool to capture intricate network patterns, achieving successes across different domains. 
However, existing GNNs require careful dataset-specific architecture designs and training from scratch on each dataset, leading to an expertise-intensive process with difficulty in generalizing across graphs from different domains. 
Therefore, it can be hard for practitioners to infer which GNN model can generalize well to other graph datasets. 
To address this challenge, we propose a novel cross-dataset pretraining framework, "one model for one graph," which overcomes the limitations of previous approaches that failed to use a single GNN to capture diverse graph patterns across datasets with significant gaps. 
Specifically, we pretrain a bank of source models, with each one corresponding to a specific dataset. When inferring to a new graph, scoring modules choose a subset of source models to effectively integrate pretraining knowledge while mitigating negative transfer. Extensive experiments consistently demonstrate the superiority of our proposed method on both link prediction and node classification tasks. 
\end{abstract}






\maketitle

\section{Introduction}

As an ubiquitous data structure, graphs can represent a wide range of structural data across different domains, such as academia~\citep{yang2016revisiting}, e-commerce~\citep{ying2018graphPINSAGE, borisyuk2024lignn, fan2019graph, tang2020knowing}, and molecule~\citep{ying2021doGraphormer}. 
Graph neural networks~(GNNs) have exhibited great performance when learning and inferring on a single graph dataset~\cite{ma2021deep}.
However, vallina GNNs often fail to generalize across graphs with node features of varying semantic meanings and dimensions.
Recently, feature heterogeneity can be mitigated by: (i) transforming node features into textual descriptions and (ii) employing Large Language Models~(LLMs) to encode them into the aligned textual representation space. 
With the help of LLM encodings, there are increasing attempts to generalize the knowledge learned by GNNs to unseen graphs, i.e., cross graph learning~\cite{oneforall,hou2024graphalign,chen2024llaga, zhao2024graphany}.





\begin{figure*}[ht]%
    \centering
    \subfloat[One model for all graphs\label{fig:existing}]{{\includegraphics[width=0.35\linewidth]{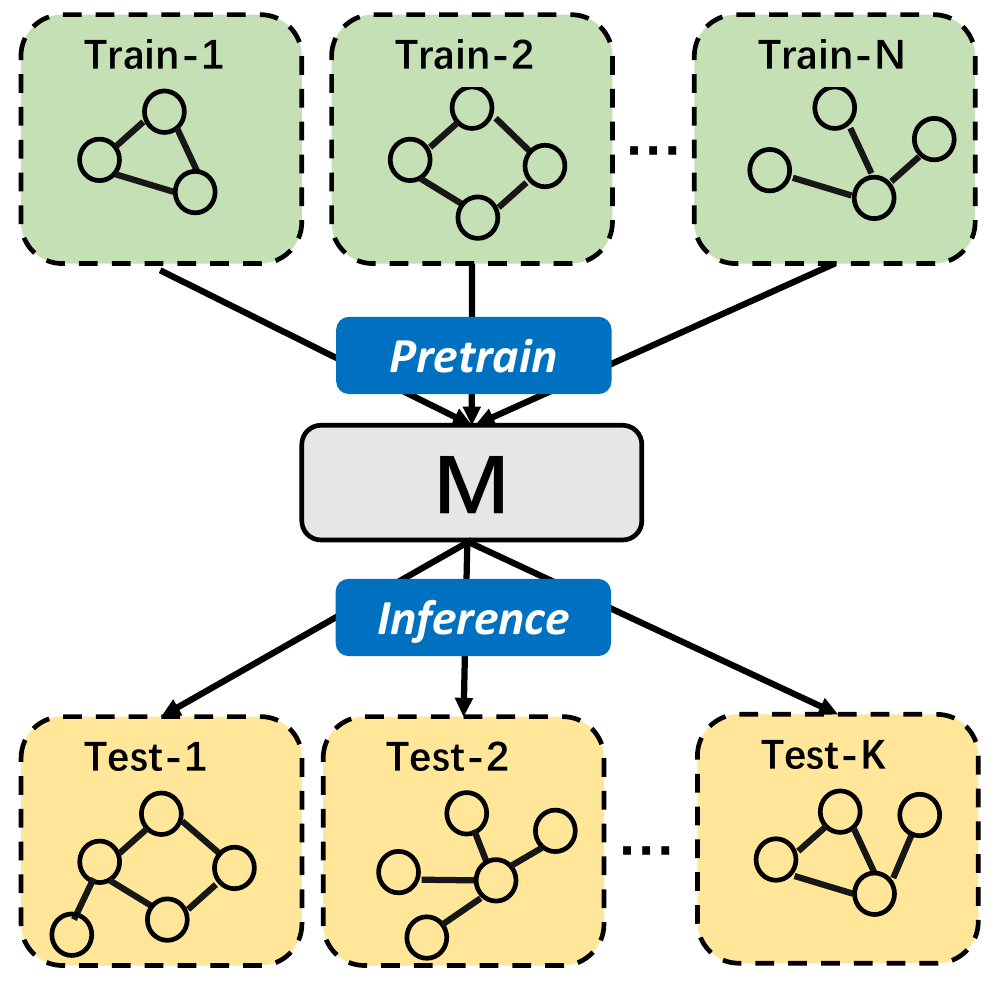}}}%
\qquad \qquad
    \subfloat[One model for one graph\label{fig:newpipeline}]{{\includegraphics[width=0.37\linewidth]{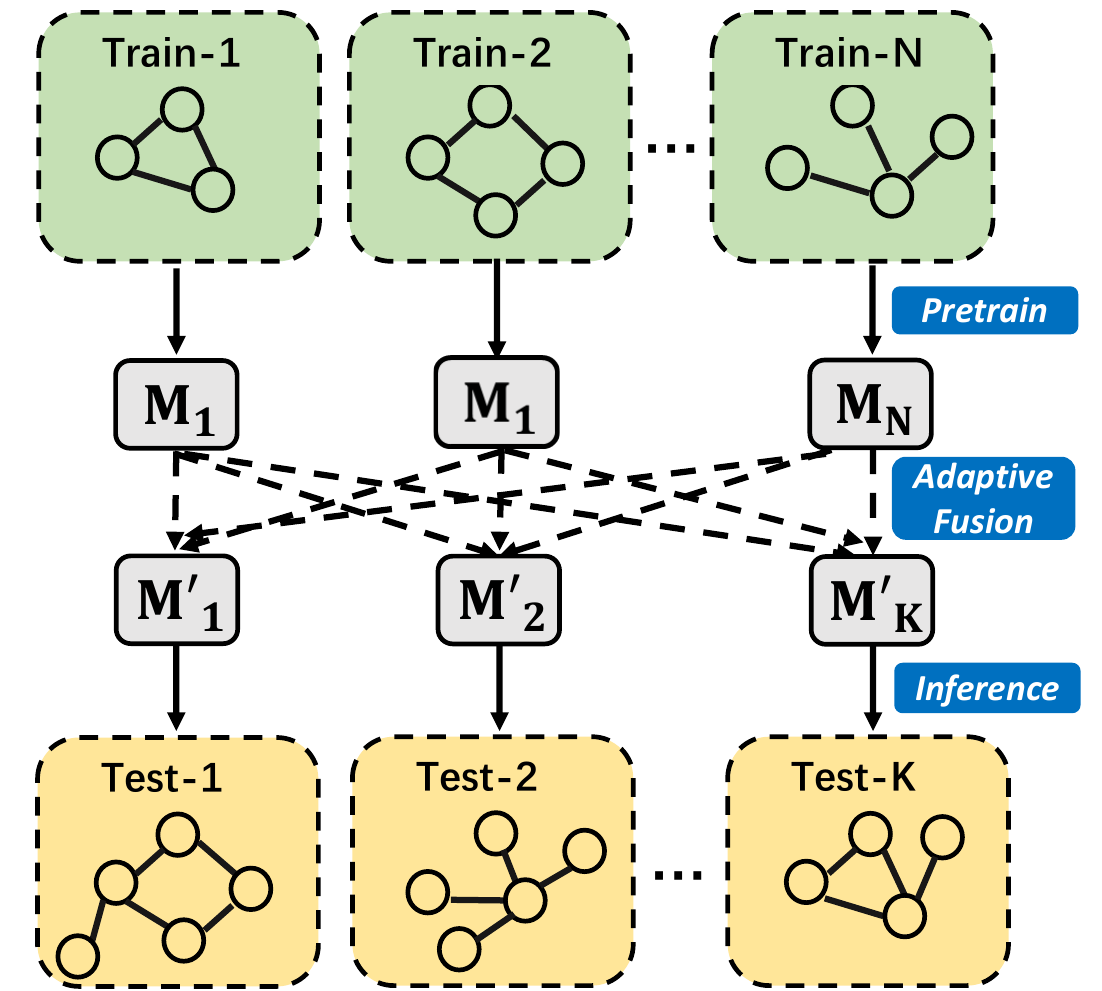} }}%
    \caption{Existing ``one model for all graphs" pipeline vs. the proposed ``one model for one graph" pipeline .}%
    \label{fig:pipelinecomparison}
    \vspace{-0.1in}
\end{figure*}

Most of the existing efforts follow a ``one model for all graphs" pipeline~\citep{oneforall, huang2023prodigy, chen2024text, chen2024llaga}. This pipeline usually pretrains only one model from all pretraining graphs and then empolys the model to all test graphs as shown in Figure~\ref{fig:existing}.
However, this paradigm has certain limitations. Graphs exhibit significantly different structural properties. For example, the homophily property, a crucial factor affecting the node classification performance of GNNs, varies significantly across graphs~\cite{ma2021homophily}. As highlighted by \cite{mao2023demystifyingNC}, a single GNN struggles to capture varying levels of homophily simultaneously. Therefore, in the pretraining stage, one single model for multiple pre-training graphs can hinder its effectiveness. Similary, in the inference stage, one pretrained model may struggle to generalize across all test graphs. A recent benchmark~\citep{chen2024text} reveals that, even within the aligned textual representation space, existing cross-graph learning methods with the ``one model for all graphs" pipeline frequently experience negative transfer from pretraining to inference. 

The aforementioned observations suggest that pretraining a single model for multiple graphs presents challenges in both the pretraining and inference stages. Therefore, in this work, we propose to individually pretrain one source model for each pretraining graph and then save the set of different source models as a model bank to effectively leverage cross-domain graphs. During inference, a subset will be automatically selected to produce a new fused model adaptive to a test graph. This proposed pipeline is different from the ``one model for all graphs" pipeline and we refer to it  as the ``one model for one graph" pipeline~(Figure~\ref{fig:newpipeline}). 
Compared to the ``one model for all graphs" pipeline, ``one model for one graph" pipeline can potentially mitigate the negative transfer in both pretraining and inferring stages. 
In the pretraining stage, the new pipeline trains one model only on one graph, which inherently reduces the feature and structural heterogeneity problems in cross-graph learning. For the inferring stage, the new pipeline will produce one fused model adaptive to a test graph, which potentially mitigates the discrepency between the training and test graphs. 
Moreover, since each model is pretrained separately, the new pipeline makes it easier to incorporate new pretraining graphs, without needing to repeat the entire pretraining process.

To enjoy the benefits of the new pipeline, we implement a novel ``one model for one graph'' pretraining framework for cross graph learning, OMOG. In OMOG, each source model consists of a set of non-parametric SGCs~\citep{wu2019simplifying} followed by an attention mechanism to capture different levels of structural information.
After pre-training the source model, OMOG fixes its parameters and trains a scoring model on the same graph, which is to score the input data by its likeness to the source data. Thus, given a set of $N$ pretraining graphs, OMOG will train a source and an associated scoring model on each graph, resulting in a bank of $N$ pretrained source and $N$ scoring models. During the inference stage, every scoring module will calculate a relevance score for a test graph, and models with high relevance scores will be fused to form a pre-trained model. Extensive experiments are conducted to demonstrate the superiority of our design in both zero-shot and few-shot learning settings.

\section{Related work}
\label{sec: rw}

The graph machine learning community has recently witnessed a growing trend to extend models designed for a single graph across different graphs (or datasets)~\citep{mao2024graphGFM}. The key obstacle to cross-graph learning stems from feature and structural heterogeneity. Early endeavors typically address feature heterogeneity by neglecting the original features~\citep{qiu2020gcc} and adopting GNN-based self-supervised learning to extract transferrable structural patterns. However, such a strategy performs poorly on text-rich networks and suffers from negative transfer~\citep{xu2023betterWITHLESS} due to the structure shift across different datasets. \citet{zhao2024allOFAAFO} adopt dimensionality reduction to unify the feature dimension while features remain poorly aligned. To generate high-quality unified features across graphs, LLM and prompt engineering\citep{oneforall} have been adopted to generate features in a unified text space~\citep{chen2024text}. \citet{oneforall} focus on the cross-data co-training where the downstream dataset aligns with the pre-training one and achieves inferior performance when transferring to novel datasets~\citep{chen2024text}. \citet{chen2024llaga, tang2023graphgpt} focus on transferring trained models to new datasets while presenting inferior performance with inadequate supervision. \citet{li2024zerog, huang2023prodigy} identify the importance of reformulating prediction into nearest neighbor retrieval for effective prediction in low supervision scenarios, while their approach which uses one dataset to fit all graphs will struggle when training across graphs and possibly suffer negative transfer. 
\citet{xia2024anygraph, hou2024graphalign} further introduce a mixture-of-expert architecture to remedy this issue, while their gating function training lacks graph-aware supervision and adopts a fixed number of experts, resulting in inferior performance. \citet{he2024unigraph} adopt a LLM-based backbone, which incurs significant computational overhead. \citet{zhao2024graphany} achieve cross-graph learning based on the label space instead of the feature space, which is orthogonal to our work. Another line of work studies cross-task learning across graphs, where \citet{jin2021automatedAUTOSSL, ju2023multiParetoGNN} focus on selecting pre-training tasks to adapt different downstream tasks, while \citet{liu2023graphpromptWWW, sun2023all, GPPT} tackle the task heterogeneity to support different tasks with a unified backbone. Our work can potentially be combined with these works to support cross-task learning across graphs.

\section{Method}
\label{sec: method}
\begin{figure*}[ht!]
    \centering
    \includegraphics[width=2\columnwidth]{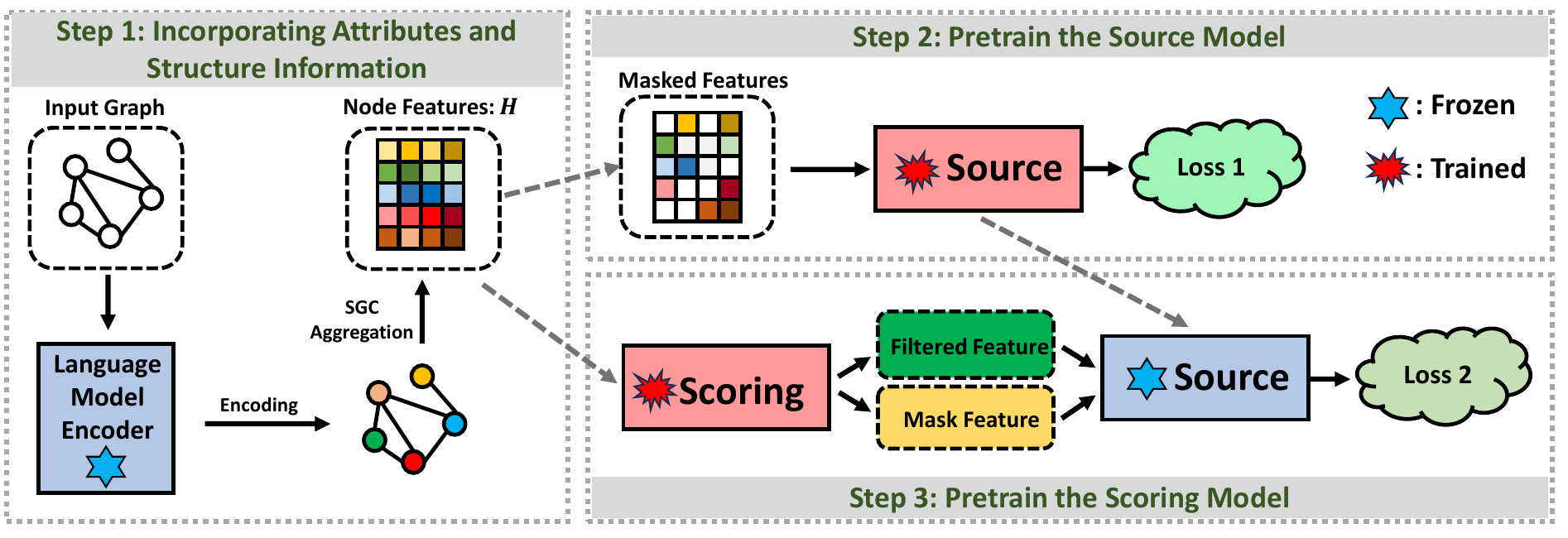}
    \caption{An illustration of the pretraining stage. The first step encodes the node attributes with language models and then applies SGC to incorporate the structure information. The second step pretrains the source model with contrastive loss. The third step trains a scoring module to filter the domain-related features.}\label{fig: model}
\end{figure*}

In this section, we introduce our one model for one graph pretraining framework with cross-domain graphs, \textbf{OMOG}. It is an implementation of the new pipeline shown in Figure~\ref{fig:newpipeline}. It consists of two stages -- the pretraining stage and the inference stage.  In the pretraining stage, OMOG will pre-train one source model for one graph with one associated scoring model separately. In the inference stage, it will adaptively choose suitable source models according to the scoring models. Before we detail these two stages, we start with introducing the problem formulation.

\textbf{Problem formulation}: In this work, we focus on text-attributed graphs~(TAGs), or more generally, text-space datasets~\citep{chen2024text} whose features can be converted into text-attributes. An input graph can be defined as $\mathcal{G}=(\mathcal{V}, \textbf{A}, \textbf{S})$, where $\mathcal{V}=\{v_1, v_2, ..., v_n\}$ is the set of $n$ nodes, and $\textbf{A} \in \mathbb{R}^{n\times n}$ represents the adjacency matrix of the graph, and $\mathbf{S}=\{\boldsymbol{\mathit{s}}_1, \boldsymbol{\mathit{s}}_2, ... \}$ is the set of text descriptions for all nodes. We focus on cross-graph pretraining with a transferring setting. Specifically, assuming that we are given $N$ pretraining graphs $\{G_{1}, \cdots, G_{N}\}$, we would like to pretrain a model bank $\mathcal{M}$ with one model for each pretraining graph and then transfer knowledge in $\mathcal{M}$ to unseen test graphs. We focus on two downstream tasks: i.e., node classification and link prediction. 
For node classification, we aim to predict the category $y_i$ of the target node $v_i$. For the link prediction, we predict whether there is a link between two target nodes $v_i$ and $v_j$.


\subsection{The Pretraining Stage}

The whole pre-training process is illustrated in Figure~\ref{fig: model},  which contains the following steps:
\begin{enumerate}[nosep, leftmargin=*]
    \item \textbf{Incorporating attribute and structure information}: To achieve cross-graph pre-training across diverse domains, we first adopt LLMs to generate node features for each graph in a unified text space. Based on the unified feature space, we adopt non-parametric message passing~\citep{wu2019simplifying} to generate node-level embeddings incorporating structural information.
    \item \textbf{Pretraining source models}: This step involves pre-training models that can effectively transfer to downstream datasets. As shown in \citep{xu2023betterWITHLESS}, pre-training a single model across graphs with diverse structural properties results in negative transfer and catastrophic forgetting. Therefore, we design a model bank to preserve pre-training knowledge. This is achieved by pre-training one separate model for each graph.
    \item \textbf{Pretraining scoring modules}: To adaptively extract the proper source models for a test graph, we pre-train scoring modules to determine the relevance between pre-trained models in the bank and test graphs. The pre-trained scoring modules can then be applied to select the most relevant source models to fuse a new model specific to test graphs in the inference stage.
\end{enumerate}

Next we introduce the technical details of these steps in the pretraining stage. 

\begin{figure*}[t]
    \centering
    \includegraphics[width=2\columnwidth]{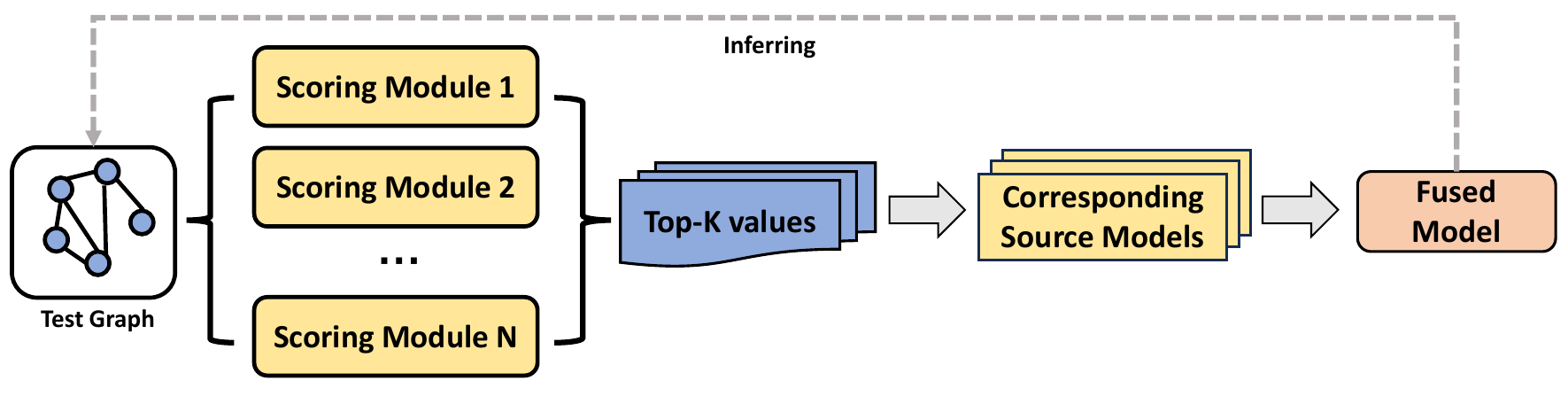}
    \caption{An illustration of the inference stage. We input the test graph features into each scoring model to calculate the relevance values to the corresponding source models. Then we select source models with top-k largest values and fuse them into a new model to infer the downstream tasks.}\label{fig: infer}
\end{figure*}




\textbf{Incorporating attribute and structure information.} A unified feature space is the requisite of cross-graph pre-training. As a result, we follow \citep{oneforall, chen2024text} to adopt LLMs as encoders to generate text embedding $\boldsymbol{\mathit{x}}_i$ based on node attributes $\boldsymbol{\mathit{s}}_i$. In this way, we get the node features $\mathbf{X} = \{\boldsymbol{\mathit{x}}_1, ..., \boldsymbol{\mathit{x}}_n\}$.
Specifically, we adopt SentenceBERT~\citep{reimers2019sentence} which exhibits promising performance in previous studies~\citep{Chen2023ExploringTP, oneforall,li2024zerog}. 

Based on the unified feature space, we subsequently apply SGC~\citep{wu2019simplifying} to integrate the graph structural information.  First, we calculate the normalized adjacency matrix of the pretraining graph,

\begin{equation}
    \mathbf{J} = \tilde{\mathbf{D}}^{-\frac{1}{2}}\tilde{\mathbf{A}}\tilde{\mathbf{D}}^{-\frac{1}{2}}
\end{equation}
where $\tilde{\mathbf{A}} = \mathbf{A}+\mathbf{I}$ is the adjacency matrix with self-loop, and $\tilde{\mathbf{D}}$ is the degree matrix of $\tilde{\mathbf{A}}$. Then we could use \textbf{J} to update the node features with neighborhood information,
\begin{equation}
    \mathbf{H}^{(\alpha)} = \mathbf{J}\mathbf{H}^{(\alpha-1)}\label{eq: sgc}
\end{equation}
where we set $\mathbf{H}^{(0)}= \mathbf{X}$, and $\alpha$ is the number of neighborhood hops that are used to update the node features.
By repeating the Equation~\ref{eq: sgc}, we can get node features $\mathbf{H}^{(\alpha)}$ integrated with different hops of structural information. 

\textbf{Pretraining source models.} As shown in \citep{xu2023betterWITHLESS}, pre-training a single model across graphs with different structural properties leads to negative transfer, primarily due to the conflicts across graphs from diverse domains. To remedy this issue, we adopt a model bank to separately pre-train each model on each graph, which stores the pre-trained knowledge in each model. Considering the heterogeneous label space across different graphs, we construct a self-supervised pretext task to pretrain source models with the learning objective adopted from \cite{yanqiao2020deep}. 

Specifically, we first augment the node-level feature 
\begin{equation}
    \boldsymbol{\mathit{h}}_i=[\boldsymbol{\mathit{h}}_i^{(0)}, \boldsymbol{\mathit{h}}_i^{(1)}, ..., \boldsymbol{\mathit{h}}_i^{(\alpha)}]
\end{equation}
by randomly masking half of the features randomly following the method in \citet{zhu2021empirical}, which results in two masked views $\boldsymbol{\hat{\mathit{h}}}_{i,0}$ and $\boldsymbol{\hat{\mathit{h}}}_{i,1}$. $\boldsymbol{\mathit{h}}_i^{(\alpha)}$ corresponds to the $i$-th row of the embedding matrix $\mathbf{H}^{(\alpha)}$. For $t$ nodes within a batch, views augmented from the same nodes are considered positive pairs, and those augmented from different nodes are considered negative pairs. To better capture the featurewise interaction, we adopt a transformer block~\citep {vaswani2017attention} as the source model backbone. The forward propagation process of a source model can thus be represented as $\boldsymbol{\hat{\mathit{f}}}_{i,j} = \mathrm{Source}(\boldsymbol{\hat{\mathit{h}}}_{i,j})$, where the full definition of $\mathrm{Source model}$ is deferred to Appendix~\ref{app: comp}. The source model is then optimized with the following contrastive loss: 
\begin{equation}
    \mathcal{L}_{1}=\sum_{i=1}^t \mathrm{log}\frac{2e^{\mathrm{sim}(\boldsymbol{\hat{\mathit{f}}}_{i,0}, \boldsymbol{\hat{\mathit{f}}}_{i,1})}}{\sum_{m=1}^t \sum_{n=1}^t [\sum_{j=0}^1 e^{\mathrm{sim}(\boldsymbol{\hat{\mathit{f}}}_{m,j}, \boldsymbol{\hat{\mathit{f}}}_{n,j})} +
    2e^{\mathrm{sim}(\boldsymbol{\hat{\mathit{f}}}_{m,0}, \boldsymbol{\hat{\mathit{f}}}_{n,1})}]
    }
\end{equation}
where $\mathrm{sim}(\cdot, \cdot)$ is the operation to calculate the cosine similarity between two vectors.

\textbf{Pretraining the scoring modules.} 
After keeping pre-trained knowledge in a model bank, we design a scoring module to decide the relevance between the corresponding source model in the bank and downstream datasets during the inference stage.
Specifically, the scoring module aims to filter key graph features related to a domain. 
Relevant features after going through the scoring model should be near the domain’s embedding cluster centroid, while unrelated features should be distant.
To achieve this goal, we need an encoder to project the sample embeddings, a filter to refine the domain-related features, and a generator to produce negative samples.
In our scenario, the well-trained source model can be reused as the encoder, thus we follow the idea in~\citep{guo2023data} and design a post-hoc scoring module.
In OMOG, we employ an MLP for both the roles of filter and generator, which will learn to generate a matrix to mask the input features and thus leave out the domain-related patterns.
Meanwhile, the mask matrix itself could be regarded as a negative sample since it is supposed to not have domain information.
We train the MLPs also in a mini-batch manner. 
When it takes in a feature  $\boldsymbol{\mathit{h}}_i$, it will generate an mask matrix by $\boldsymbol{\mathit{a}}_i = \mathrm{MLP}(\boldsymbol{\mathit{h}}_i)$.
Then the filtered feature is calculated as $\boldsymbol{\tilde{\mathit{h}}}_i = \boldsymbol{\mathit{h}}_i+\boldsymbol{\mathit{a}}_i$, which is viewed as a positive sample of the domain.
Meanwhile, the mask matrix $\boldsymbol{\mathit{a}}_i$ is viewed as a negative sample.
Then we will use the source model to encode $\boldsymbol{\tilde{\mathit{h}}}_i$ and $\boldsymbol{\mathit{a}}_i$ repectively, it will result in a positive embedding $\boldsymbol{\tilde{\mathit{f}}}_i=\mathrm{Source}(\boldsymbol{\tilde{\mathit{h}}}_i)$ and a negative embedding $\boldsymbol{\mathit{o}}_i=\mathrm{Source}(\boldsymbol{\mathit{a}}_i)$.
We want the positive embedding to be close to the centroid of the domain embedding cluster while the negative embedding to be distant from the centroid $\boldsymbol{\mathit{f}}_{center}$, that the training loss of the scoring is designed as below,
\begin{equation}
    \mathcal{L}_{2}=\mathrm{dis}(\boldsymbol{\tilde{\mathit{f}}}_i,\boldsymbol{\mathit{f}}_{center})+\frac{1}{\mathrm{dis}(\boldsymbol{\mathit{o}}_i, \boldsymbol{\mathit{f}}_{center})}
\end{equation}

where $\mathrm{dis}(\cdot, \cdot)$ is the Euclidian distance between two vectors, and $\boldsymbol{\mathit{f}}_{center}$ can be calculated as $\boldsymbol{\mathit{f}}_{center}=\mathrm{MEAN}(\mathrm{Source}(\mathbf{H}))$.

\begin{table*}[htbp]
\centering
\caption{The transferring comparison under the zero-shot setting. Note that ``NC'' refers to node classification; ``LP'' refers to link prediction; and ``Rank" is calculated based on the average rank of each model on each dataset. For results of more baselines, please refer to the appendix.}
\label{tab: zero}
\resizebox{\linewidth}{!}{
\begin{tabular}{@{}l|l|lcccccccc|c}
\toprule
\textbf{Task} &
\textbf{Methods} &

 \textbf{Child} &
 \textbf{History} &
\textbf{Cora} &
\textbf{Citeseer} &
\textbf{Dblp} &
\textbf{Products} &
\textbf{Pubmed}&
\textbf{Sports} &
 \textbf{Wikics} 
 & \textbf{Rank}\\ \midrule
\multirow{5}{*}{\textbf{NC}}& \textbf{Oneforall}      & 12.56 & 13.54 & 34.29 & 39.66 & 46.81 & 13.45 & 35.73 & 11.05 & 40.26       & 4.67\\
&\textbf{LLaGA}                                       & 13.75 & 14.58 & 33.78 & 40.79 & 47.53 & 17.26 & 35.38 & 12.35 & 39.37      & 4.22\\
&\textbf{AnyGraph}                                    & 13.84 & 15.16 & 55.63 & 40.03 & 50.27 & 22.36 & 37.92 & 15.35 & 50.84      & 3.00\\
&\textbf{ZeroG}                                       & \underline{18.41} & \underline{21.88} & \underline{60.43} & \underline{42.65} & \underline{52.81} & \underline{25.89} & \textbf{41.75} & \underline{18.97} & \underline{57.26}      & 1.89\\ \cmidrule{2-12}
&\textbf{OMOG}                                        & \textbf{20.34} & \textbf{25.68} & \textbf{66.19} & \textbf{49.23} & \textbf{57.53} & \textbf{31.02} & \underline{39.71} & \textbf{23.65} & \textbf{62.42 } 
     & \textbf{1.11}\\ \midrule
\multirow{5}{*}{\textbf{LP}}&\textbf{Oneforall}   
                                   & 15.28 & 10.83 & 17.46 & 16.52 & 13.31 & 13.77 & 15.35 & 14.30 & 15.83       & 4.78\\
&\textbf{LLaGA}                    & 14.65 & 16.21 & 18.01 & 19.66 & 17.43 & 12.64 & 17.81 & 15.87 & 21.27      & 4.22\\
&\textbf{AnyGraph}       &\underline{26.24} & \underline{28.63} & \underline{54.24} & \underline{47.94} & \underline{49.64} & \underline{33.76} & \underline{46.93} & 32.59 & \underline{49.82}      & 2.11\\
&\textbf{ZeroG}                    & 21.83 & 24.39 & 49.36 & 43.18 & 41.08 & 31.27 & 40.28 & \underline{33.98} & 45.19    & 2.89\\ \cmidrule{2-12}
&\textbf{OMOG}&       \textbf{31.29} & \textbf{34.86} & \textbf{56.28} & \textbf{50.72} & \textbf{53.46} & \textbf{40.95} & \textbf{49.42} & \textbf{37.81} &
\textbf{52.38} 
     & \textbf{1.00}\\ \bottomrule
    
\end{tabular}}
\end{table*}

\subsection{The Inference Stage}\label{sec: inference}
After pretraining a bank of source and scoring models, we can adopt them to infer the unseen test data as shown in Figure~\ref{fig: infer}. 
Similar to the forward propagation process of pretraining the scoring models, the feature will first be filtered by the scoring models and then encoded by the source models. 
Finally, the cosine similarity between the output and the centroid embedding of the domain will be calculated as a relevance score to indicate how likely the sample is related to the domain.
For a test graph $\mathcal{G}_{test}$, we first get the node embeddings $\mathbf{H}_{test}$=[$\mathbf{H}_{test}^{(0)}$, ..., $\mathbf{H}_{test}^{(\alpha)}$] aggregated by SGC.
Subsequently, we will feed $\mathbf{H}_{test}$ into every scoring model to compute its domain-related representations.
For the $p$\textsuperscript{th} source and scoring model which is trained on graph $\mathcal{G}_p$ with node embeddings as $\mathbf{H}_p$, the relevance score is calculated as follows:
\begin{equation}
    v_p = \mathrm{sim}(\mathrm{MEAN}(\mathrm{Source}(\mathrm{Scoring}(\boldsymbol{\mathit{h}}_{test}))), \boldsymbol{\mathit{f}}_{center,p}))
\end{equation}
where the $\mathrm{sim}(\cdot, \cdot)$ is the operation to calculate the cosine similarity between two vectors, and 
$\boldsymbol{\mathit{f}}_{center,p}$ can be calculated as $\boldsymbol{\mathit{f}}_{center,p}=\mathrm{MEAN}(\mathrm{Source}(\mathbf{H}_p))$.

After getting the relevance values for all source models, we will select top-k values with $\mathcal{E}=\mathrm{top{-}k}(v_1, v_2, ...)$. Then we scale the weights with $\mathrm{softmax}(\mathcal{E})$.
Next, we would use them to weight their corresponding source models to produce a fused model.
The fusion process merges the parameters of multiple source models by calculating their weighted average. It can be viewed as an example of weighted model merging~\cite{matena2022merging},
which can serve as a powerful tool to mitigate the negative transfer~\cite{li2024tackling}.
Specifically, by adaptively fusing models trained on diverse graphs, our approach enables better generalization across datasets, preserving beneficial test-graph-aware knowledge while mitigating harmful influences. 
This approach not only improves adaptability in graph heterogeneity but also enhances the efficiency of incorporating new pertaining datasets.
Moreover, we prove the following the theorems below for the ability of OMOG fusing mechanism, for which the detailed proof can be found in Appendix~\ref{app:theory}.

\vspace{1mm}
\noindent \textbf{Theorem 1. \textit{OMOG’s merging strategy is theoretically equivalent to Bayesian Model Averaging.}}
\vspace{1mm}

\noindent \textbf{Theorem 2. \textit{OMOG’s Top-K selection and merging guarantee that the fused model outperforms or matches the best individual
source model in expectation}}
\vspace{1mm}

Once the fused model is ready, we can use it to infer the target node feature $\boldsymbol{\mathit{h}}_{test}$ and generate the output embeddings $\boldsymbol{\mathit{f}}_{test}$.
For zero-shot node classification, the label whose embedding has the highest cosine similarity with the test node output embedding is regarded as the prediction.
For zero-shot link prediction, the logit of link existence is the cosine similarity between the two test nodes' output embeddings.






\textbf{Extension to few-shot learning setting.} 
For the few-shot learning, we follow the same process as zero-shot learning to produce a pretrained model. The key difference from zero-shot node classification is that we use both the label embedding and the centroid embedding of each class to compute the final predictions. 
Specifically, suppose that there are $s$ classes in the support sets, we input all the samples in the support set and calculate the average of output embeddings for each class.
Thus, for each class, there is a centroid embedding $\boldsymbol{\mathit{f}}_{avg, i}$, where $0\leq i\leq s$.
Suppose that the label embedding of each class is $\boldsymbol{\mathit{l}}_{i}$, then the predicted label $y_{test}$ for a test node with output embedding $\boldsymbol{\mathit{f}}_{test}$ can be calculated as following,
\begin{equation}
    y_{test} = \mathrm{argmax}_{0\leq i\leq s}[\mathrm{sim}(\boldsymbol{\mathit{f}}_{test},\boldsymbol{\mathit{f}}_{avg, i})+\mathrm{sim}(\boldsymbol{\mathit{f}}_{test},\boldsymbol{\mathit{l}}_{i})]
\end{equation}
where the $\mathrm{sim}(\cdot, \cdot)$ is the operation to calculate the cosine similarity between two vectors.
\section{Experiment}
In this section, we conduct comprehensive experiments to evaluate the effectiveness of our proposed method OMOG from the following perspectives:
\begin{enumerate}[nosep, leftmargin=*]
    \item \textbf{RQ1:} Can our method effectively transfer pre-trained models to unseen test data in zero-shot and few-shot settings?
    \item \textbf{RQ2:} How does each component of our method influence the transfer effectiveness?
    \item \textbf{RQ3:} Why does source model selection notably enhance transfer effectiveness?
\end{enumerate}


\begin{table*}[htbp]
\centering
\caption{The results for few-shot node classification. Note that ``Rank" is calculated based on the average rank of each model on each graph dataset.}
\label{tab:few nc}
\resizebox{\linewidth}{!}{
\begin{tabular}{@{}l|lccccccccc|c}
\toprule
\textbf{Methods} &

 \textbf{Ratings} &
 \textbf{Child} &
 \textbf{History} &
\textbf{Cora} &
\textbf{Citeseer} &
\textbf{Dblp} &
\textbf{Products} &
\textbf{Pubmed}&
\textbf{Sports} &
 \textbf{Wikics} & \textbf{Rank} 
\\ \midrule
\textbf{GCC}               & 23.25 & 17.86 & 18.14 & 33.28 & 35.62 & 34.52 & 21.04 & 35.11            & 16.48 & 29.93 & 7.7      \\
\textbf{GraphMAE}          & 22.68 & 18.74 & 19.94 & 35.79 & 37.20 & 38.18 & 20.87 & 36.34            & 18.42 & 28.87  & 7.3    \\
\textbf{Oneforall}         & 26.73 & 27.81 & 26.59 & 56.26 & 40.27 & 46.24 & 31.27 & 39.93            & 23.91 & 41.74  & 5.8    \\
\textbf{LLaGA}             & 31.51 & 29.26 & 27.28 & 53.23 & 42.15 & 43.28 & 32.86            & 40.27 & 25.22 & 43.37  & 5.1   \\
\textbf{GraphAlign}        & 34.79 & 32.69 & 32.71 & 72.86 & 52.39 & 58.60 & 44.62 & \textbf{50.76}   & 32.65 & 63.17   & 3.4  \\
\textbf{GCOPE}             & \underline{37.85} & 32.73 & 36.29 & 72.17 & 55.87 & 60.24 & \underline{46.02} & 48.10            & \underline{35.88} & 59.28   & 2.8  \\
\textbf{Prodigy}           & 30.88 & \underline{33.63} & \underline{35.82} & \textbf{77.59} & \underline{56.28} & \underline{60.83} & 45.35& 44.87    & 33.18 & \underline{64.23} & 1.9   \\ \midrule
\textbf{OMOG}              & \textbf{39.23} & \textbf{35.87} & \textbf{38.25} & 75.41 & \textbf{59.36} & \textbf{63.24} & \textbf{46.27} & \underline{49.82}            & \textbf{36.72} & \textbf{65.39}  & \textbf{1.1}
    \\ \bottomrule
\end{tabular}}
\end{table*}

\subsection{Experimental Setup}

\textbf{Datasets.} We utilize $10$ diverse texture-attributed graphs sourced from \citet{chen2024text}. These datasets span a wide range of domains, including citation networks, social networks, and e-commerce networks. The graph sizes range from thousands to millions of nodes, with the number of classes across datasets spanning from $3$ to $39$. These datasets exhibit both domain shift and structural property shift~\citep{chen2024text}, effectively reflecting the challenges encountered when transferring pre-trained graph models to novel domains in real-world scenarios. For a comprehensive overview of the datasets, please refer to Appendix~\ref{app:dataset}.


\textbf{Evaluation settings.} To evaluate the effectiveness of our methods under the transferring setting~(\textbf{RQ1}), we use a widely adopted setting that pre-trained models that are adapted to unseen test datasets with little (few-shot) or no downstream task supervision (zero-shot)~\citep{chen2024text, oneforall}. We consider both node classification and link prediction as target tasks. To test the transferring capability of models, we adopt a \textit{leave-one-out} strategy. Specifically, given $10$ datasets, each time, one of them will be selected as the target downstream test data, and the other nine datasets will be used as the pre-training data. Regarding the evaluation metrics, we utilize accuracy for node classification and Hits@100 for link prediction. 

\textbf{Baselines.} To demonstrate the effectiveness of our framework, we consider state-of-the-art cross-graph learning baselines, which can be categorized as the ``single model'' and ``mixture of models'' frameworks. 
The\textit{``Single model''} framework adopts a unified backbone to achieve cross-graph learning, with representatives including OneForAll~\citep{oneforall}, GCOPE~\citep{zhao2024allinone}, LLaGA~\citep{chen2024llaga}, ZeroG~\citep{li2024zerog} and Prodigy~\citep{huang2023prodigy}. Their major difference lies in the selection of backbone models, where OneForAll, ZeroG, GCOPE, and Prodigy are still based on GNNs, while LLaGA adopts LLM. Specifically, Prodigy and ZeroG transform the prediction into a nearest neighbor retrieval problem.  Graph self-supervised learning baselines, including GCC~\citep{qiu2020gcc} and GraphMAE~\citep{hou2022graphmae}, also belong to this category. 
The \textit{``mixture of models''} framework adopts a group of models to be pre-trained and then transferred to downstream tasks. Two representative methods include AnyGraph~\citep{xia2024anygraph} and GraphAlign~\citep{hou2024graphalign}. They directly apply the MoE architecture~\citep{shazeer2017outrageouslyMOE} without the correspondence between each model and each graph. Additionally, they utilize a fixed number of expert models, essentially functioning as a ``multiple models for multiple graphs" approach. The implementation details of our method can be found in Appendix~\ref{app:implementation}.

\begin{figure}[ht!]
    \centering
    \subfloat{
    \begin{minipage}[b]{0.24\textwidth}
    \includegraphics[width=\columnwidth]{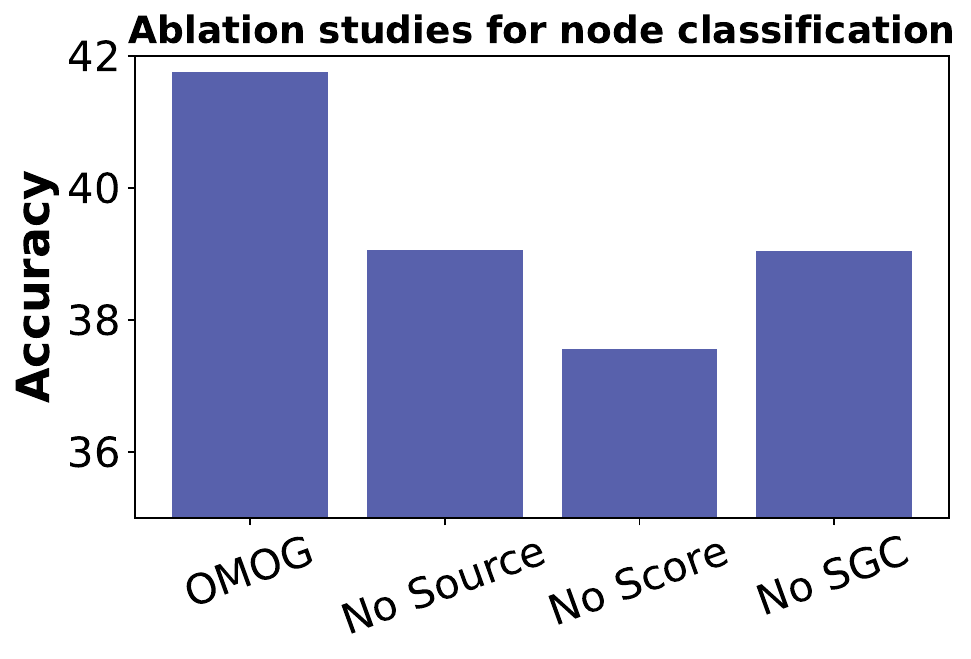}
    \end{minipage}
    }
    \subfloat{
    \begin{minipage}[b]{0.24\textwidth}
    \includegraphics[width=\columnwidth]{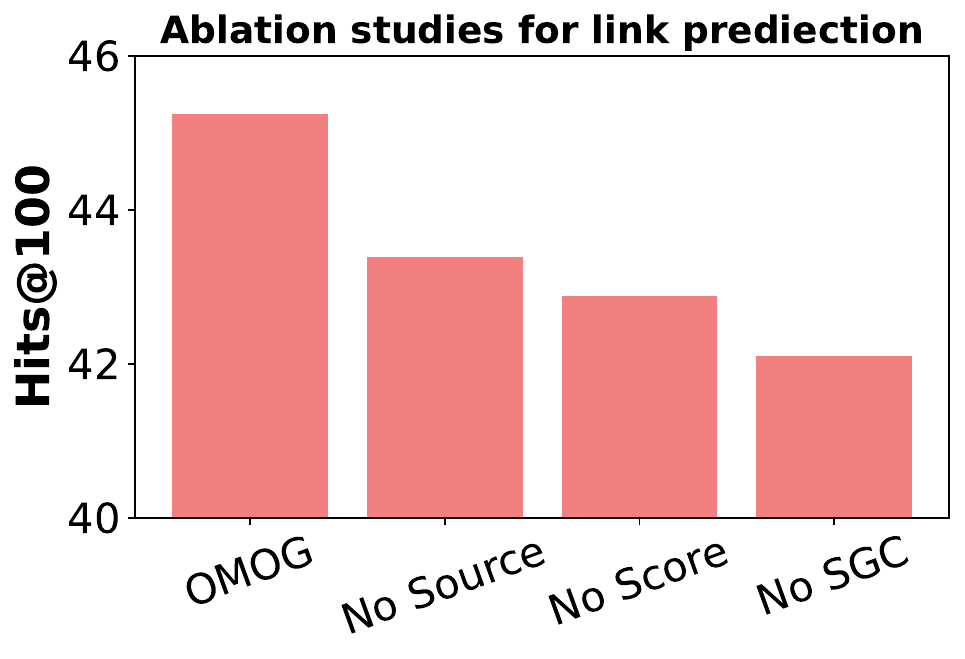}
    \end{minipage}
    }
    \vspace{-1em}
    \caption{The impact for key components on OMOG}\label{fig: alation}
\end{figure}

\begin{figure}[ht!]
    \centering
    \subfloat{
    \begin{minipage}[b]{0.24\textwidth}
    \includegraphics[width=\columnwidth]{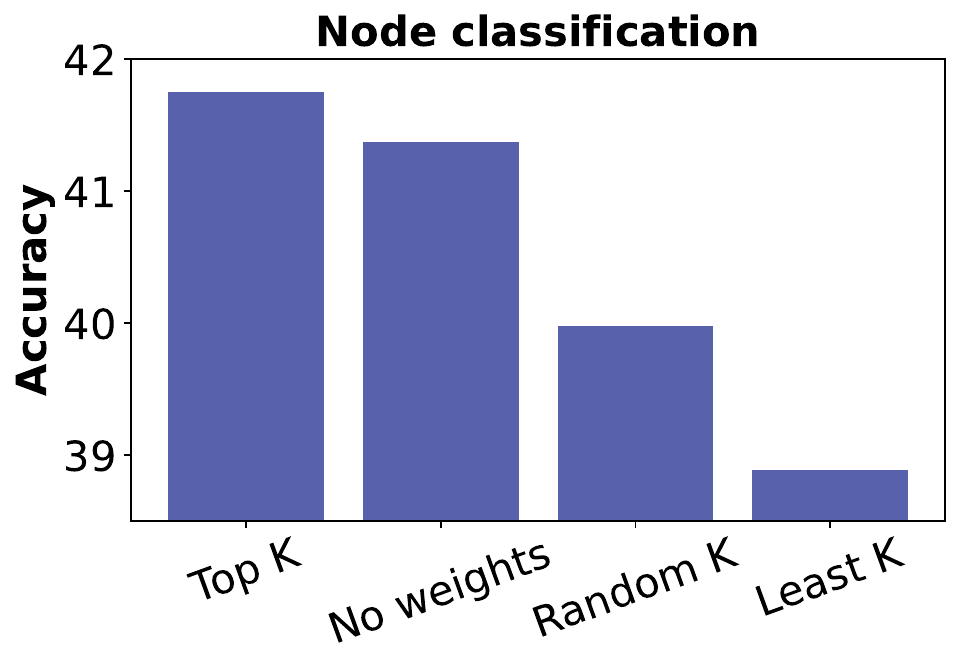}
    \end{minipage}
    }
    \subfloat{
    \begin{minipage}[b]{0.24\textwidth}
    \includegraphics[width=\columnwidth]{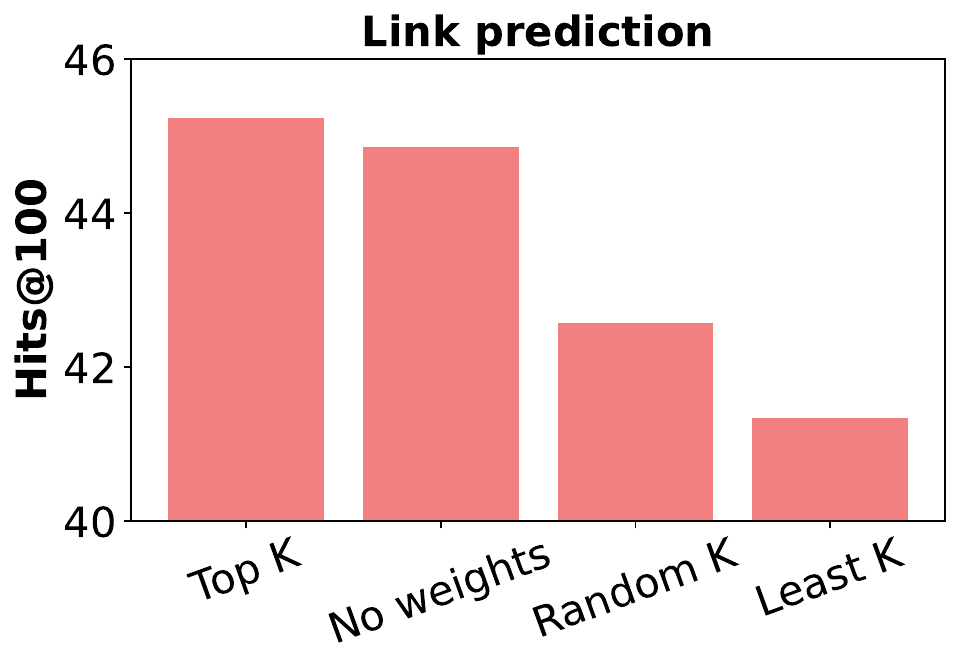}
    \end{minipage}
    }
    \vspace{-1em}
    \caption{The performance of different scoring module designs.}\label{fig: gate more}
\end{figure}

\subsection{RQ1: Evaluating the transferability }

In this subection, we evaluate the transferability of different cross-graph pretraining methods by comparing their performance on downstream tasks. Specifically, we focus on zero-shot and few-shot settings.

\subsubsection{Transferring in a zero-shot setting}

We first evaluate different cross-graph pretraining methods under zero-shot learning scenarios. We choose all baseline models applicable to the zero-shot learning settings, including OneForAll, LLaGA, AnyGraph, ZeroG, and our method. Since LLaGA adopts an LLM as the backbone model, it takes a considerably longer time to evaluate using the \textit{leave-one-out} strategy. As a result, we pre-train it on Arxiv and Products and test it on every downstream task. To prevent data leakage when the target dataset is Product, we pre-train it using Arxiv and Sports. The results are shown in Table~\ref{tab: zero}. 

From Table~\ref{tab: zero}, we make the following observations:
\begin{itemize}[nosep, leftmargin=*]
    \item \textbf{Our method consistently performs better in node classification and link prediction tasks.} Our method achieves the best performance on $8$ out of $9$ datasets for the node classification task and all of the datasets for the link prediction task, demonstrating that our method achieves superior transferability. In node classification and link prediction, our method outperforms the second-best baselines by a margin of 9\%. Moreover, our method requires substantially less computation time than baselines like ZeroG, which requires fine-tuning the LLM. 
    \item \textbf{Viewing zero-shot prediction as nearest neighbor retrieval is critical for effective zero-shot prediction.} Comparing the performance of each baseline, we find that the performance of OneForAll and LLaGA is consistently lower than other baselines. The key difference between these methods lies in their inference strategy. Specifically, OneForAll and LLaGA directly make inferences based on the model classification head, while other baselines project target embeddings and label embeddings into the same space for nearest neighbor retrieval. 
    \item \textbf{Vanilla mixture-of-model can not solve the data heterogeneity problem effectively.} Despite adopting a mixture of model architecture, our model outperforms AnyGraph by a large margin, especially in node classification, which archives a 20\% improvement on average. Comparing the design of these two models, our model presents two key distinctions: (1) we adopt one model for one graph; and (2) we adopt an adaptive set of models when transferring to downstream tasks. These two designs make our models better tackle heterogeneity when transferring and achieve better performance. 
\end{itemize}

\subsubsection{Transferring in a few-shot setting}

We then evaluate different cross-graph pretraining methods under the few-shot setting. We consider all applicable baselines, including GCC, GraphMAE, OneForAll, LLaGA, GraphAlign, GCOPE, and Prodigy. For OneForAll and Prodigy, we follow \citep{oneforall} to augment the target prediction subgraph with graph prompts sampled from each class. GraphAlign and our method use the inference strategy introduced in Section~\ref{sec: inference} to generate label embeddings for each class. Other baseline methods directly adopt the few-shot labels as supervision to fine-tune the prediction head. For Prodigy, we follow the original setting to pretrain the model on a subset of the MAG240M dataset. Considering that most baseline methods are designed for node classification, we present the results for few-shot node classification. For each dataset, we randomly select $5$ samples for each class.

As shown in Table~\ref{tab:few nc}, we summarize the main observations below: 
\begin{itemize}[nosep, leftmargin=*]
    \item \textbf{Our method outperforms other baseline methods}. Our method performs best on $8$ of the $10$ downstream datasets. Comparing our method to the best baseline Prodigy, our method significantly outperforms it on heterophilous dataset, i.e., Ratings. Our method achieves more than 5\% improvement compared to Prodigy despite using less pre-training data. This demonstrates our method's transferability to unseen downstream datasets with different structural properties. 
    \item \textbf{Our method demonstrates more superiority on datasets with complicated label space.} Only on Cora and Pubmed, whose class numbers are 3 and 5, respectively, the performance GraphAlign can slightly surpass our method. For more complicated cases where the class number of the dataset is more than $10$, our method consistently outperforms other baselines. Compared to GraphAlign, which uses an ``mixture of models'' design, our method achieves an improvement of over 6\% on average. 
\end{itemize}





\subsection{RQ2: Ablation Study}

We then study how each key component of OMOG affects the transferring effectiveness to answer \textbf{RQ2}. We identify three key components of OMOG:
\begin{itemize}[nosep, leftmargin=*]
    \item \textbf{Source model} acts as the backbone model to solve the prediction task for each graph. 
    \item \textbf{SGC module} generates node-level graph embeddings using message passing.
    \item \textbf{Scoring model} takes node-level embeddings as input and generates a relevance score to select the related source models.
\end{itemize}

As shown in Figure~\ref{fig: alation}, we find that 
\begin{enumerate}[nosep, leftmargin=*]
    \item \textbf{Every component contributes to effective transferring.} This ablation study reveals that each component significantly contributes to the model's overall performance. 
    \item \textbf{Scoring mechanism is crucial to cross-graph node classification.} For node classification, we find that removing the scoring mechanism results in a significant performance drop, which suggests that scoring plays an important role in addressing data heterogeneity by adaptively selecting source models from the proper domain.
    As a comparison, SGC components play the most important role in link prediction, which means structural information is vital for link prediction.
    \item \textbf{LLM embedding plays an important role.} When solely using the aggregated LLM embedding for prediction, the model can still have good performance, indicating the importance of aligned feature space in the zero-shot learning scenario.
\end{enumerate}

\begin{figure}[ht!]
    \centering
    \subfloat{
    \begin{minipage}[b]{0.24\textwidth}
    \includegraphics[width=\columnwidth]{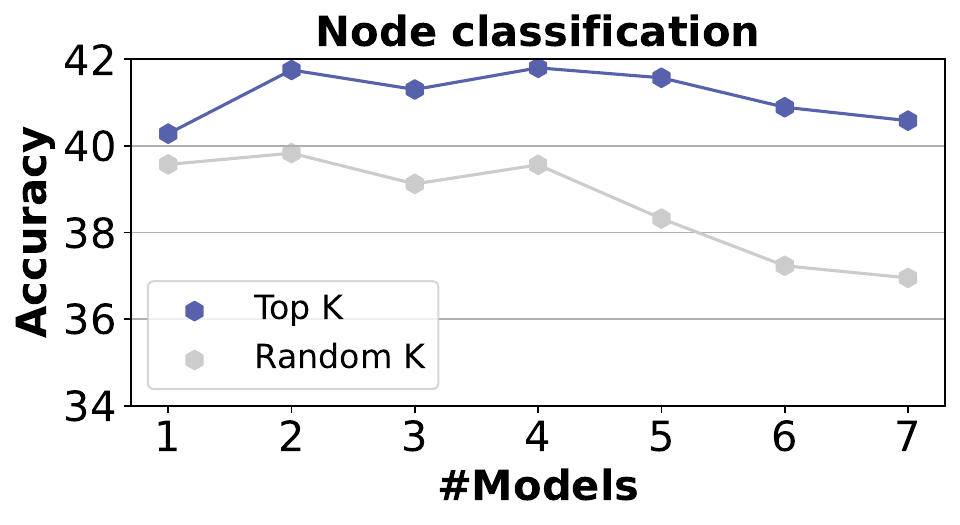}
    \end{minipage}
    }
    \subfloat{
    \begin{minipage}[b]{0.24\textwidth}
    \includegraphics[width=\columnwidth]{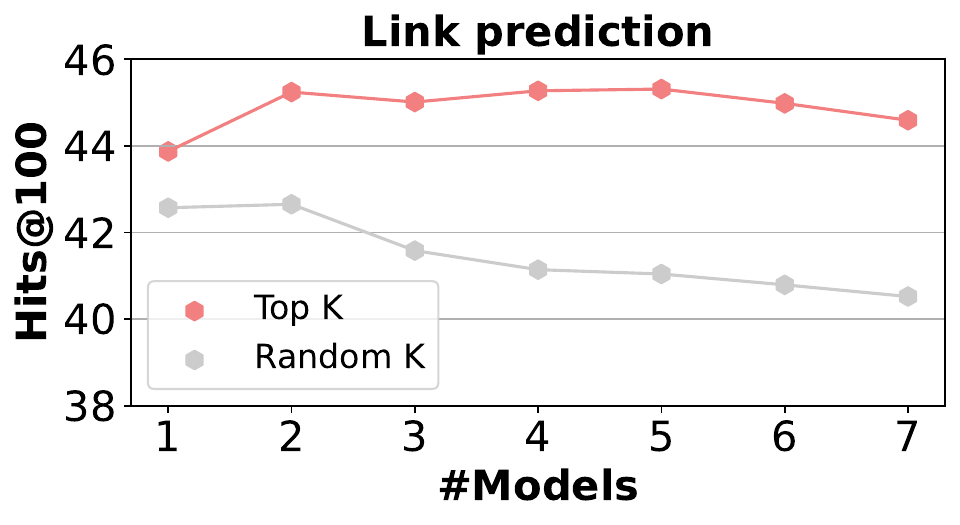}
    \end{minipage}
    }
    \vspace{-1em}
    \caption{The effect of the number of source models.}\label{fig: gate num}
\end{figure}

\subsection{RQ3: Investigating the Scoring Module Design}

Considering the importance of source model selection and distinguishing our work from existing ones~\citep {hou2024graphalign, xia2024anygraph}, we further study the influence of different source model selection strategies to answer \textbf{RQ3}. We compare the following strategy variants to our original design stated in Section~\ref{sec: method}: (1) ``No weights'' still adopts the TopK selection strategy while removing the weights for each source model; (2) ``Random K'' randomly selects $K$ source models instead of models with highest scores; and (3) ``Least K'' selects $K$ source models with lowest scores. 

As shown in Figure~\ref{fig: gate more}, the scoring model design in OMOG outperforms all variants, which suggests that 
\begin{itemize}[nosep, leftmargin=*]
    \item The score learned by our scoring module can guide us to select the most helpful source models for transferring, and consequently making the ``TopK'' strategy outperforms the ``Random K'' and ``Least K'' strategies.
    \item The weight given by the scoring module can further help the fusing of selected source models, which makes the ``TopK'' strategy outperforms the ``No weights'' strategy. 
\end{itemize}

Furthermore, we check how the number of selected source models $K$ affects the performance, and the results are shown in Figure~\ref{fig: gate num}: with the increase of the number of source models, the performance of ``Top K" first increases and then becomes relatively stable while the performance of ``Random K"  will consistently decrease, indicating significant negative transfer. This observation supports that the scoring mechanism in our design can help mitigate the negative transfer when including more pretraining graphs.

\subsection{Computation Complexity and Training Time}
Here we compare the efficiency of OMOG against the baselines. For a single expert model, the training time complexity is $O(d^2+K^2d)$, where $K$ is the number of hops aggregated for a target node and $d$ is the dimension of the feature vector. For a single gate model, the training time complexity is also $O(d^2+K^2d)$. Thus, the pretraining time complexity on a single dataset is just $O(d^2+K^2d)$. Moreover, since our framework allows to train mutiple experts and gates on different datasets in parallel, it actually would gain efficiency advantages compared to the previous cross-domain pretraining methods. This is also proved by the empirical observation in Table~\ref{tab:runtime_gpu}, which shows OMOG consumes less time and GPU memory during training.

\begin{table}[h]
\centering
\caption{Comparison of training time and GPU memory usage.}\label{tab:runtime_gpu}
\resizebox{\linewidth}{!}{
\begin{tabular}{lccccccc}
\toprule
 & GraphMAE & Oneforall & LLaGA & AnyGraph & GraphAlign & ZeroG & OMOG \\
\midrule
Time (h)       & 3.7 & 12.3 & 20.3 & 9.8 & 3.8 & 21.6 & 0.9 \\
GPU Memory (G) & 8.3 & 17.0 & 35.6 & 14.5 & 6.7 & 39.6 & 4.1 \\
\bottomrule
\end{tabular}}

\end{table}

\section{Conclusion}

In this paper, we present a new perspective together with an easy yet effective framework, ``one model for one graph'' (\textbf{OMOG}), to achieve effective cross-graph learning. Through extensive experiments, we develop the following practices for cross-graph learning: training one source model for each graph and then utilizing scoring functions to select the source models most proper for downstream tasks adaptively. Our perspective can benefit future development in related areas, such as graph foundation models.


\appendix

\section{Dataset Details \label{app:dataset}}
For the detailed statistics of the datasets we use in the experimets, we record them in Table~\ref{tab:stat}.

\begin{table}[h]
\centering
\caption{Details of our datasets.}
\label{tab:stat}
\resizebox{\linewidth}{!}{
\begin{tabular}{@{}llllllll@{}}
\toprule
\textbf{Name} & \textbf{\#Graphs} & \textbf{\#Nodes} & \textbf{\#Edges} & \textbf{Domains} & \textbf{Tasks} & \textbf{\#Classes} & \textbf{Metrics} \\ \midrule
\textbf{Cora}           & 1      & 2708   & 10556    & CS Citation  & Node, Link & 7    & Accuracy, Hits@100            \\
\textbf{CiteSeer}       & 1      & 3186   & 8450     & CS Citation  & Node, Link & 6    & Accuracy, Hits@100            \\
\textbf{Arxiv}          & 1      & 169343 & 2315598  & CS Citation  & Node, Link & 40   & Accuracy, Hits@100            \\

\textbf{History}        & 1      & 41551  & 503180   & E-commerce   & Node, Link & 12   & Accuracy, Hits@100            \\
\textbf{Child}          & 1      & 76875  & 2325044  & E-commerce   & Node, Link & 24   & Accuracy, Hits@100            \\
\textbf{Sportsfit}      & 1      & 173055 & 3020134  & E-commerce   & Node, Link & 13   & Accuracy, Hits@100            \\
\textbf{Products}       & 1      & 316513 & 19337722 & E-commerce   & Node, Link & 39   & Accuracy, Hits@100            \\
\textbf{Amazon Ratings} & 1      & 24492  & 186100   & E-commerce   & Node, Link & 5    & Accuracy, Hits@100            \\
\textbf{Pubmed}         & 1      & 19717  & 88648    & Bio Citation & Node, Link & 3    & Accuracy, Hits@100            \\
\textbf{WikiCS}         & 1      & 11701  & 431726   & Knowledge    & Node, Link & 10   & Accuracy, Hits@100            \\

\textbf{DBLP(*)}           & 1      & 14376  & 431326   & CS Citation  & Node, Link & 4    & Accuracy, Hits@100                         
                      \\ \bottomrule
\end{tabular}}
\end{table}
\section{Computation}
\label{app: comp}

The computation formula for the source model is shown as follow:
Given input $H$ with shape $(B, \alpha, d)$, where $B$ represents the batch size, $\alpha$ represents the number of $\alpha$ SGC heads, and $d$ represents the hidden dimension. 
Then, $H_{o} = \text{softmax}\left(\frac{HW_q(W_k^TH^T)}{\sqrt{d_k}}\right)HW_v$, 
$H_{1} = \text{LayerNorm}(H + H_{o})$, and finally $\hat{\mathit{f}}= \text{LayerNorm}(H_1 + \text{MLP}(H_1))$. 

\section{Implementations \label{app:implementation}}
In this section, we present our detailed implementations of OMOG. 
For the vector length of language embeddings, we set them to 384 to balance the efficiency and performance.
For the number $\alpha$ of SGC operations, we set it to 4.
And we choose top-2 models in the fusion stage to select 2 source models which largest relevance scores with the downstream task.

In the pretraining stage for source and scoring models, we use Adam~\citep{kingma2014adam}. 
The initial learning rate is set to be 0.0001.
\newtheorem{assumption}[theorem]{Assumption}
\section{Theoretical Framework}\label{app:theory}
\noindent\textbf{Notations and Definition}

\begin{definition}[Graph Domain and Graph Representations]
A \textbf{graph domain} \( \mathcal{G} \) is defined as a structured dataset:
\[
\mathcal{G} = (V, A, S, Y),
\]
where:
\begin{itemize}
    \item \( V \) is the set of nodes,
    \item \( A \) is the adjacency matrix,
    \item \( X \) is the set of node attributes,
    \item \( Y \) is the set of labels (possibly missing in unsupervised cases).
\end{itemize}
Each graph \( G_i \) in a collection of source graphs \( \{\mathcal{G}_i\}_{i=1}^{N} \) has an associated pre-trained model \( M_i \) trained specifically on that graph.
\end{definition}

\begin{definition}[Relevance Score and KL Divergence Approximation]
For a test graph \( G_{\text{test}} \), the \textbf{relevance score} \( v_i \) of a source model \( M_i \) is computed as:
\[
v_i = \text{sim}(\text{MEAN}(\text{Source}(\text{Scoring}(H_{\text{test}}))), f_{\text{center}, i}),
\]
where \( f_{\text{center}, i} \) is the feature center of source model \( M_i \).

We assume that \( v_i \) serves as an \textbf{approximate inverse measure} of the KL divergence between the output distributions:
\[
v_i \approx - D_{KL}( P(y|x, G_{\text{test}}) || P(y|x, G_i) ).
\]
\end{definition}

\begin{assumption}[Negative Transfer and Structural Mismatch]
Let \( M_{\text{joint}} \) be a single model pre-trained on multiple graphs simultaneously. Due to \textbf{structural mismatches across graphs}, negative transfer occurs, leading to degraded performance:
\[
\mathcal{L}(M_{\text{joint}}, G_{\text{test}}) > \mathcal{L}(M_i, G_{\text{test}}), \quad \forall i \in \text{relevant source models}.
\]
Thus, training \textbf{separate} source models \( M_i \) preserves graph-specific knowledge.
\end{assumption}

\begin{assumption}[Gaussian Output Distributions in Graph Domains]
We assume that the predictive distributions \( P(y|x, G) \) for different graphs are approximately Gaussian:
\[
P(y|x, G) \sim \mathcal{N}(\mu_G, \sigma_G^2).
\]
For any two graphs \( G_{\text{test}} \) and \( G_i \), their respective output distributions are:
\[
P(y|x, G_{\text{test}}) \sim \mathcal{N}(\mu_{\text{test}}, \sigma_{\text{test}}^2), \quad P(y|x, G_i) \sim \mathcal{N}(\mu_i, \sigma_i^2).
\]
This assumption follows standard Bayesian learning principles, where uncertainty in predictions is modeled via Gaussian distributions.
\end{assumption}

\noindent\textbf{Key lemma and propostion}

\begin{lemma}[KL Divergence Approximation in Gaussian Case]
Under \textbf{Assumption 2}, the KL divergence between the predictive distributions of the test graph \( G_{\text{test}} \) and a source graph \( G_i \) is given by:
\[
D_{KL}(P(y|x, G_{\text{test}}) \parallel P(y|x, G_i)) = \frac{1}{2} \left( \frac{\sigma_i^2}{\sigma_{\text{test}}^2} - 1 - \log \frac{\sigma_i^2}{\sigma_{\text{test}}^2} \right).
\]
For small differences in variance (\( |\sigma_i^2 - \sigma_{\text{test}}^2| \ll \sigma_{\text{test}}^2 \)), using the first-order Taylor expansion of \( \log(1+x) \), we obtain the approximation:
\[
D_{KL}(P(y|x, G_{\text{test}}) \parallel P(y|x, G_i)) \approx \frac{1}{2} (\sigma_i^2 - \sigma_{\text{test}}^2).
\]
Since \( \sigma_{\text{test}}^2 \) is constant across models, the OMOG similarity score \( v_i \) can be approximated as:
\[
v_i \approx - \frac{1}{2} \sigma_i^2.
\]
Thus, the similarity score \( v_i \) serves as an \textbf{inverse proxy} for the predictive variance \( \sigma_i^2 \), which aligns with Bayesian Model Averaging (BMA) theory.
\end{lemma}
\textbf{Proof:} 

By the definition of KL divergence for two Gaussian distributions \( \mathcal{N}(\mu_{\text{test}}, \sigma_{\text{test}}^2) \) and \( \mathcal{N}(\mu_i, \sigma_i^2) \), with the assumption that \( \mu_{\text{test}} = \mu_i \) (mean alignment assumption), the KL divergence simplifies to:

\[
D_{KL}(P_{\text{test}} \parallel P_i) = \frac{1}{2} \left( \frac{\sigma_i^2}{\sigma_{\text{test}}^2} - 1 - \log \frac{\sigma_i^2}{\sigma_{\text{test}}^2} \right).
\]

For small differences in variance (\( |\sigma_i^2 - \sigma_{\text{test}}^2| \ll \sigma_{\text{test}}^2 \)), we use the first-order Taylor expansion \( \log(1+x) \approx x \) when \( x \approx 0 \):

\[
\log \frac{\sigma_i^2}{\sigma_{\text{test}}^2} \approx \frac{\sigma_i^2}{\sigma_{\text{test}}^2} - 1.
\]

Substituting into the KL divergence expression:

\[
D_{KL}(P_{\text{test}} \parallel P_i) \approx \frac{1}{2} (\sigma_i^2 - \sigma_{\text{test}}^2).
\]

Since \( \sigma_{\text{test}}^2 \) is constant across models, the OMOG similarity score \( v_i \) can be approximated as:

\[
v_i \approx - \frac{1}{2} \sigma_i^2.
\]

Thus, the similarity score \( v_i \) serves as an \textbf{inverse proxy} for the predictive variance \( \sigma_i^2 \), aligning with Bayesian Model Averaging (BMA) theory.
\qed
\begin{proposition}[Optimal Model Selection with Bayesian Model Averaging]
Given \( v_i \approx -\frac{1}{2} \sigma_i^2 \), the \textbf{OMOG model selection weight} \( w_i \) computed via softmax:
\[
w_i = \frac{\exp(v_i)}{\sum_{j=1}^{N} \exp(v_j)}
\]
approximates the \textbf{Bayesian Model Averaging (BMA) weight}:
\[
w_i^{\text{BMA}} = \frac{1/\sigma_i^2}{\sum_{j=1}^{N} 1/\sigma_j^2}.
\]
Thus, \textbf{OMOG’s merging strategy is theoretically equivalent to Bayesian Model Averaging}, ensuring an optimal combination of source models.
\end{proposition}
\textbf{Proof:} 

Substituting \( v_i \approx -\frac{1}{2} \sigma_i^2 \) into the softmax function:

\[
w_i = \frac{\exp(-\frac{1}{2} \sigma_i^2)}{\sum_{j=1}^{N} \exp(-\frac{1}{2} \sigma_j^2)}.
\]

Using the approximation \( \exp(-x) \approx 1/x \) for small \( x \):

\[
w_i \approx \frac{1/\sigma_i^2}{\sum_{j=1}^{N} 1/\sigma_j^2}.
\]

This is exactly the Bayesian Model Averaging weight, proving that OMOG’s merging strategy is theoretically equivalent to BMA.
\qed

\noindent\textbf{Main Theorem and Corollary}

\begin{theorem}[OMOG Model Fusion Minimizes Expected Transfer Error]
Let \( M_{\text{fused}} \) be the model obtained by merging source models:
\[
M_{\text{fused}} = \sum_{i \in \mathcal{E}} w_i M_i, \quad \text{where } \mathcal{E} = \text{Top-K}(v_i).
\]
where \( w_i \) is the adaptive weight derived from softmax over \( v_i \). The expected error on the test domain satisfies:
\[
\mathbb{E}[(f_{\text{fused}}(x) - f^*(x))^2] \leq \min_{i} \mathbb{E}[(f_i(x) - f^*(x))^2].
\]
Thus, OMOG’s \textbf{Top-K selection and merging guarantee that the fused model outperforms or matches the best individual source model} in expectation.
\end{theorem}
\textbf{Proof:}

The expected squared error can be decomposed using the standard bias-variance decomposition:
\[
\mathbb{E}[(f(x) - f^*(x))^2] = \underbrace{(\mathbb{E}[f(x)] - f^*(x))^2}_{\text{Bias}^2} + \underbrace{\text{Var}(f(x))}_{\text{Variance}}.
\]

From Bayesian Model Averaging (BMA), the fused model’s expectation is given by:
\[
\mathbb{E}[f_{\text{fused}}(x)] = \sum_{i \in \mathcal{E}} w_i \mathbb{E}[f_i(x)],
\]
and its variance satisfies:
\[
\text{Var}(f_{\text{fused}}(x)) = \sum_{i \in \mathcal{E}} w_i^2 \text{Var}(f_i(x)).
\]

Since OMOG assigns model weights as \( w_i \propto \frac{1}{\sigma_i^2} \), models with lower predictive variance receive higher weights. Consequently, the total variance satisfies:
\[
\text{Var}(f_{\text{fused}}(x)) \leq \min_i \text{Var}(f_i(x)).
\]

Since the bias of a weighted ensemble is at most the minimum bias among the base models, we assume:
\[
\mathbb{E}[f_{\text{fused}}(x)] \approx \mathbb{E}[f_{\text{best}}(x)].
\]

Therefore, we compare the expected squared error:
\[
\mathbb{E}[(f_{\text{fused}}(x) - f^*(x))^2] = \underbrace{(\mathbb{E}[f_{\text{fused}}(x)] - f^*(x))^2}_{\text{Bias}^2} + \underbrace{\text{Var}(f_{\text{fused}}(x))}_{\text{Variance}}.
\]

Since \(\text{Var}(f_{\text{fused}}(x))\) is minimized and \(\text{Bias}^2\) does not increase significantly due to Top-K selection, we conclude:
\[
\mathbb{E}[(f_{\text{fused}}(x) - f^*(x))^2] \leq \min_{i} \mathbb{E}[(f_i(x) - f^*(x))^2].
\]

Thus, OMOG’s fusion strategy ensures that the fused model achieves at least the best individual model’s performance in expectation.
\qed
\begin{corollary}[Mitigation of Negative Transfer]
Since OMOG \textbf{only selects source models with high \( v_i \)} while suppressing those with low scores, it inherently reduces \textbf{negative transfer} compared to joint training:
\[
\mathcal{L}(M_{\text{fused}}, G_{\text{test}}) < \mathcal{L}(M_{\text{joint}}, G_{\text{test}}).
\]
\end{corollary}

\section*{GenAI Usage Disclosure}
In any stage of this research works, we did not use any generative AI tools.

\bibliographystyle{ACM-Reference-Format}
\bibliography{main}
\end{document}